\documentclass[10pt, a4paper]{article}
\usepackage{lrec}
\usepackage{multibib}
\newcites{languageresource}{Language Resources}
\usepackage{multirow}
\usepackage{graphicx}
\usepackage{tabularx}
\usepackage{soul}
\usepackage{hyperref}
\usepackage{breakurl}

\usepackage{epstopdf}
\usepackage[latin1]{inputenc}

\usepackage{hyperref}
\usepackage{xstring}

\title{MMCR4NLP: Multilingual Multiway Corpora Repository for Natural Language Processing}

\name{Raj Dabre, Sadao Kurohashi}

\address{Kyoto University, Kyoto University \\
         Kyoto, Japan, Kyoto, Japan \\
         prajdabre@gmail,com, kuro@i.kyoto-u.ac.jp\\
         }

\abstract{
Multilinguality is gradually becoming ubiquitous in the sense that more and more researchers have successfully shown that using additional languages help improve the results in many Natural Language Processing tasks. Multilingual Multiway Corpora (MMC) contain the same sentence in multiple languages. Such corpora have been primarily used for Multi-Source and Pivot Language Machine Translation but are also useful for developing multilingual sequence taggers by transfer learning. While these corpora are available, they are not organized for multilingual experiments and researchers need to write boilerplate code every time they want to use said corpora. Moreover, because there is no official MMC collection it becomes difficult to compare against existing approaches. As such we present our work on creating a unified and systematically organized repository of MMC spanning a large number of languages. We also provide training, development and test splits for corpora where official splits are unavailable. We hope that this will help speed up the pace of multilingual NLP research and ensure that NLP researchers obtain results that are more trustable since they can be compared easily. We indicate corpora sources, extraction procedures if any and relevant statistics. We also make our collection public for research purposes.\\ \newline  }

\begin{document}

\maketitleabstract

\section{Introduction}
Text Corpora form the backbone of data-driven Natural Language Processing tasks ranging from automatic text segmentation to syntactic and semantic analysis to discourse. Bilingual parallel corpora which contain the same sentences in two languages are not only useful for Machine Translation tasks but also enable one to use an analysis tool developed for one language for another language by using transfer learning. Multilingual Multiway corpora (also known as N lingual corpora; terms which we will use interchangeably) are special corpora where the same sentence is available in multiple languages. Formally speaking a N-way corpus is one in which the same sentence is present in N languages. 

Large trilingual corpora are quite common since most countries maintain transcripts of various meetings (business, legal etc.) in English, the native language and an additional language. This additional language can depend on the situation such as geographical proximity or diplomatic and economic relations. The ASPEC corpus which is a trilingual Japanese-Chinese-English corpus is a product of joint collaboration between Japan and China in order to boost relations and promote research.

One of the most attractive features of a N-lingual corpus is that by adding an additional language it automatically gets direct links to each of the N languages. This is a very desirable property since it is possible to perform transfer learning from one language to another. For example it should be possible to transfer part of speech tagging or parsing information from a language which has annotated data and parsers to a language that has none.

In this paper we will list out various N-lingual corpora which are either publicly available or have been extracted by us. We list the extraction procedures along with various corpora level statistics. We also group corpora by language families where ever possible so that it becomes easier to study linguistic phenomena for related languages. We make our collection public so that people can work on them directly instead of having to spend time on searching for and extracting them.

Most of these corpora are available online but they are not organized for multilingual experiments and researchers end up spending a significant amount of time writing boilerplate code to organize and use said corpora. Moreover, because there is no official MMC collection researchers tend to use their own splits of the datasets which makes it difficult to compare against their proposed approaches. And thus in the cases where official development and test sets are unavailable we have created our own training, development and test splits which we hope will be used by everyone to ensure fair comparison of methodologies and their results. Our key contributions are as follows:
\begin{itemize}
\item We have systematically collected, organized (or identified) and made available N lingual corpora from Europarl, TED talks, ILCI, Bible and UN corpora.
\item We have also made available some of the scripts we used to organize our corpus collection, thereby eliminating the need to write boilerplate code.
\item Our collection spans 5 domains and can be used for studies on domain adaptation and transfer learning.
\item We have also organized some of the corpora by grouping languages accoring to language families to facilitate NLP for related languages and investigate how language relatedness impacts various transfer learning tasks.
\end{itemize}

%Similarly the Europarl corpus spans 21 languages spoken in the European union.

\section{Related Work}
Our work revolves around accumulating and organizing existing N-lingual corpora. The most popular examples are: United Nations \cite{ZIEMSKI16.1195}, Europarl \cite{koehn2005epc}, Ted Talks \cite{cettoloEtAl:EAMT2012}, ILCI \cite{JHA10.874} and Bible \cite{Christodouloupoulos2015} corpora. These corpora have been used mainly for machine translation \cite{zoph-knight:2016:N16-1,och2001statistical} and for various studies on language relatedness studies \cite{W16-1208}, cross lingual part of speech tagging \cite{P15-2044} and cross lingual parsing \cite{DBLP:journals/tacl/AgicJPASS16}. To the best of our knowledge there has been no active work on creating a single compilation of multilingual corpora.

\section{Extraction Procedures and Languages}
\subsection{Corpus Extraction Mechanism}
Most corpora are not directly available as N-way corpora but as N-1 bilingual corpora where one of the languages is always English. As such we simply extract the N-way corpus by retaining the N-1 sentences which have the same English translation. We use the following procedure:

\begin{itemize}
    \item target-sentences = hashmap(hashset())
    \item all-corpora = hashmap(hashmap())
    \item for each language-pair, corpus in corpora:
    \begin{itemize}
    \item for source-sentence, target-sentence in corpus:
    \begin{itemize}
        \item target-sentences[language-pair].add(target-sentence)
        \item all-corpora[language-pair][target-sentence] = source-sentence
    \end{itemize}
    \end{itemize}
    \item common-sentences = intersect([target-sentences[key] for key in target-sentences.keys()])
    \item for sentence in common-sentences:
    \begin{itemize}
        \item write sentence to corresponding file
    
        \item for language-pair in language-pair-list: \begin{itemize}
        \item source-sentence = all-corpora[language-pair][sentence]
        \item write source-sentence to corresponding file
        \end{itemize}
        
    \end{itemize}
\end{itemize}

We load all the corpora in a dictionary to ensure quick extraction. Although it might seem to quite memory intensive our method works well in practice since there are not too many corpora that are large enough to cause out of memory issues. Additionally we will make the scripts, for extracting the N lingual corpora from text and XML files, publicly available.

\subsection{Languages}\label{sec:languages}
Our N-lingual corpus collection includes the following 59 languages: Afrikaans (afr), Albanian (al), Arabic (ar), Bulgarian (bg), Cebuano (ce), Chinese (zh), Creole (cre), Croatian (cr), Czech (cz), Danish (da), English (en), Esperanto (esp), Estonian (et), Farsi (fa), Finnish (fi), French (fr), German (de), Greek (gr), Hebrew (he), Hindi (hi), Hungarian (hu), Icelandic (ic), Indonesian (id), Italian (it), Japanese (ja), Kannada (kn), Korean (ko), Latin (la), Latvian (lt), Lithuanian (li), Malagasy (mg), Malayalam (ma), Maori (mao), Marathi (mr), Myanmar (my), Nepali (ne), Norwegian (no), Paite (pa), Polish (po), Portuguese (pt), Punjabi (pu), Qeqchi (qe), Romanian (ro), Russian (ru), Serbian (se), Slovak (sl), Slovene (sv), Somali (so), Spanish (es), Swedish (sw), Tagalog (tg), Taiwanese (tw), Tamil (ta), Telugu (te), Thai (th), Turkish (Tu), Vietnamese (Vi), Xhosa (Xh), Zarma (Za). For the remainder of the paper we will use these codes for brevity. 

\section{Multilingual Multiway Corpora}
In this section we list all the multilingual multiway corpora we have managed to acquire/identify along with relevant statistics. Our collection is available here: \url{http://lotus.kuee.kyoto-u.ac.jp/~raj/mmcr4nlp/}\footnote{In this folder we include readme files for each corpus collection for additional details which are not included in this paper due to the limited number of pages.}

\subsection{UN corpus}
The UN corpus\footnote{https://conferences.unite.un.org/uncorpus} spans 6 languages (ar, fr, ru, en, zh and es) is directly available in its N-lingual form. It contains roughly 11.36M lines and the average sentence lengths vary from 23 for Arabic to 30 for Spanish. Additionally there are 6 lingual development and test sets of 4K lines each. Since the UN corpus is directly available in its 6 lingual form we do not include it in our (downloadable) collection.

\subsection{Spoken Language and Subtitles corpus}
The Spoken Language and Subtitles corpus is an excellent source of parallel sentences in the spoken language domain. We have three different sources of TED talks corpora, two\footnote{We believe that previous IWSLT tasks showcase bilingual corpora that can be converted into N lingual versions but we leave this for future work.} of which come from the IWSLT 2016 and 2017 shared tasks and the third which was crawled from the TED website\footnote{https://www.ted.com/talk?page=N (where N is an integer greater than 0)}.

\subsubsection{IWSLT 2016 corpus}
\begin{table}[t]
\setlength{\tabcolsep}{.2em}
\small
\begin{center}
\begin{tabular}{|l|l|l|l|l|}
\hline
\textbf{Type} & \textbf{Languages}          & \textbf{train}  & \textbf{dev2010} & \textbf{tst2010/tst2013} \\ \hline
\textbf{3 lingual}   & Fr, De, En         & 191381 & 880     & 1060/886        \\ \hline
\textbf{4 lingual}   & Fr, De, Ar, En     & 84301  & 880     & 1059/708        \\ \hline
\textbf{5 lingual}   & Fr, De, Ar, Cs, En & 45684  & 461     & 1016/643        \\ \hline
\end{tabular}
\end{center}
\caption{Statistics for the the N-lingual corpora extracted from the IWSLT 2016 corpus for the languages French (Fr), German (De), Arabic (Ar), Czech (Cs) and English (En)}
\label{tab:iwslt2016}
\end{table}
The IWSLT 2016 task focused on 5 languages (fr, de, ar, cz and en) where English was the target language. One development set (dev2010) and 4 test sets (tst2010 to tst2013) are available. This corpus is not directly 5 lingual and thus using English as the pivot we extracted 3, 4 and 5 lingual versions\footnote{\url{http://lotus.kuee.kyoto-u.ac.jp/~raj/mmcr4nlp/iwslt2016/}} of the parallel corpus. We found that the test sets for 2011 and 2012 are not 5 lingual and thus exclude them from our collection. Refer to table~\ref{tab:iwslt2016} for details. The 5 lingual corpus is extremely small (~45k lines) and is a good candidate for low reosource multilingual experiments. The 3 lingual (fr, de and en) version of the corpus is almost 4 times larger indicating low overlap of the English sentences for Arabic-English and Czech-English with the French-English and German-English corpora.
\subsubsection{IWSLT 2017 corpus}
\begin{table}[t]
\setlength{\tabcolsep}{.4em}
\small
\begin{center}
\begin{tabular}{|l|l|l|l|l|}
\hline
\textbf{Type} & \textbf{Languages}          & \textbf{train}  & \textbf{dev2010} & \textbf{tst2010} \\ \hline
\textbf{3 lingual}   & De, Nl, En         & 172735 & 715     & 1363        \\ \hline
\textbf{5 lingual}   & De, Nl, It, Ro, En     & 145105  & 627     & 1153        \\ \hline
\end{tabular}
\end{center}
\caption{Statistics for the the N-lingual corpora extracted from the IWSLT 2017 corpus for the languages Dutch (Nl), German (De), Italian (It), Romanian (Ro) and English (En)}
\label{tab:iwslt2017}
\end{table}
The IWSLT 2017 task focused on 5 languages (de, nl, it, ro and en) but the objective was on a single multilingual system. One development set (dev2010) and one test set (tst2010) were provided. Like the IWSLT 2016 corpus this corpus is not directly 5-lingual either. We extracted 3 lingual (de, nl and en) and 5 lingual (de, nl, it, ro and en) versions\footnote{\url{http://lotus.kuee.kyoto-u.ac.jp/~raj/mmcr4nlp/iwslt2017/}} of the corpus. We do not list the 4 lingual version since it is of roughly the same as the 5 lingual corpus. Refer to table~\ref{tab:iwslt2017} for details.
\subsubsection{Generic Ted Talks Corpus}
We found an unofficial 13 lingual (ar, de, es, fr, he, it, ja, ko, nl, pt (Brazilian Portuguese), ru, zh (Mainland Chinese) and tw (Taiwanese Chinese)) TED talks corpus of 349049 lines which was crawled\footnote{https://github.com/ajinkyakulkarni14/TED-Multilingual-Parallel-Corpus} from the TED talks site. This repository also contains many pairs of bilingual corpora but one unusual aspect of this corpus is that it does not contain English as either a source or a target language. A 4 lingual version of this corpus which spans only ja, ko zh and tw contains an additional 40K lines for a total of 389764 lines. Since, there is no specific development or test set in the case of this 13 lingual corpus\footnote{\url{http://lotus.kuee.kyoto-u.ac.jp/~raj/mmcr4nlp/ted}} we created our own splits. For both these 4 and 13 lingual corpora we remove the last 4000 sentences (from the end of the corpus) and split them into development and test sets of 2000 sentences each. We believe that this corpus should be useful for future IWSLT tasks which focus on multilinguality.

\subsection{Bible corpus}
The Bible corpus is probably the only corpus which is translated into over 100 languages. However the corpus available online is present in XML format and needs to be preprocessed. We developed a simple XML parsing script that can produce a N-lingual version given the XML files for each language. While inspecting the corpus we discovered that the bible is not fully translated into about 40 of the languages and thus we exclude them from our collections. The English version of the bible contains about 31102 verses and we only considered the languages which contain 30000 or more translated verses. Another problem is that some verses (each of which have a unique id in the XML file) of the bible are not available in some of the XML files leading to fewer number of N lingual entries. 

In order to make it easier for researchers to work on languages belonging to the same language family we extracted N lingual corpora for the following 8 language families: Slavic, Uralic, Indo Aryan, Dravidian, Germanic, East Asian, South East Asian and Romance. For all these corpora groups we remove the last 2000 sentences (from the end of the corpus) and split them into development and test sets of 1000 sentences each. We extract the following N lingual versions\footnote{\url{http://lotus.kuee.kyoto-u.ac.jp/~raj/mmcr4nlp/bible}} of the corpus where English is always one of the languages:
\begin{itemize}
    \item 55 lingual spanning all the languages mentioned in Section~\ref{sec:languages} except Punjabi, Tamil, Taiwanese and Latvian. We chose these 55 languages since these are the only ones that are completely translated (with the exception of a few accidental omissions). This 55 lingual corpus contains 26121 lines and the missing 5000 lines are a result of the randomly missing translations for a number of verses.
    \item 9 lingual Slavic languages corpus of 30350 lines which includes bg, cr, cz, en, po, ru, se, sl and sv.
    \item 8 lingual Romance languages corpus of 30133 lines which includes en, esp, fr, it, la, pt, ro and es.
    \item 5 lingual Indo Aryan languages corpus of 30049 lines which includes en, hi, mr, my and ne.
    \item 8 lingual Germanic languages corpus of 28854 lines which includes af, da, nl, en, de, ic, no and sw.
    \item 4 lingual Dravidian languages corpus of 30651 lines which includes en, kn, ml and te.
    \item 4 lingual East Asian languages corpus of 31063 lines which includes en, zh, ja and ko.
    \item 7 lingual South-east Asian languages corpus of 29621 lines which includes en, ce, id, ma, ta, th and vi.
    \item 4 lingual Uralic languages corpus of 30885 lines which includes en, et, fi and hu.
\end{itemize}

\subsection{ILCI corpus}
The ILCI corpus has been used frequently in the Indian Languages Machine Translation shared tasks in ICON 2014\footnote{http://ltrc.iiit.ac.in/icon/2014} and 2015\footnote{http://ltrc.iiit.ac.in/icon2015/}. The ILCI corpus is a 6 lingual multilingual corpus spanning the languages Hindi, English, Tamil, Telugu, Marathi and Bengali was provided as a part of the task. The training, development and test sets\footnote{\url{http://lotus.kuee.kyoto-u.ac.jp/~raj/mmcr4nlp/ilci}} contain 45600, 1000 and 2400 6-lingual sentences respectively. Half of the corpus (train/dev/test) belongs to the tourism domain and the other half to the health domain. There is work on a 12 lingual equivalent of the ILCI corpus\footnote{http://sanskrit.jnu.ac.in/ilci/index.jsp} but it is not publicly available. Most of the sentences in the corpus are short and simple in terms of grammatical complexity. The average sentence length varies from 12 for Tamil (morphologically rich) to 17 for English (morphologically poor) indicating that most of the sentences are intended to be used in survival situations. This is quite different from the case of the UN corpus where the average sentence length for English is around 25.

\subsection{Europarl corpus}
The Europarl corpus covers the following 21 European languages: bg, cz, da, de, gr, en, es, et, fi, fr, hu, it, li, lt, nl, po, pt, sl, sv, and sw. Since the corpus available online\footnote{http://www.statmt.org/europarl/} is not directly N lingual but is available as XX-English bilingual pairs we used English to extract a 21 lingual corpus. One issue with extracting a large 21 lingual corpus is that for 11 out of the 21 languages only 30\% of the full corpus is available. Countries such as Bulgaria and Romania joined the European Union in 2007 and hence the Europarl corpus has about 400K sentences for Bulgarian and Romanian as compared to about 2M sentences for English and Spanish, languages which have been present since the beginning.

Just like we did for the Bible corpus, we also extract N lingual corpora for the following language families: Germanic, Slavic, Uralic, Baltic and Romance. Since English is the pivot language we used for extraction we include it in all the collections. As in the case of the TED corpus we remove the last 4000 sentences (from the end of the corpus for each group) and split them into development and test sets of 2000 sentences each. The details of the corpora\footnote{\url{http://lotus.kuee.kyoto-u.ac.jp/~raj/mmcr4nlp/europarl}} are as follows:

\begin{itemize}
    \item 21 lingual spanning all the languages. This corpus is of 189310 lines.
    \item 10 lingual spanning en, da, de, es, fi, fr, it, nl, pt and sv. These 10 languages are the largest in the collection. This corpus is of 1.071M lines.
    \item 6 lingual Slavic languages corpus of 342845 lines which includes bg, cz, en, po, sl and sv. We also extract a 4 lingual Slavic languages corpus of 569962 lines which includes sl, sv, cz and en. This increase in the number of lines is due to the exclusion of Polish and Bulgarian which contain significantly fewer number of entries in their respective monolingual corpora.
    \item 6 lingual Romance languages corpus of 294192 lines which includes en, fr, it, pt, ro and es. We also extract a 5 lingual version of 1.454M lines by excluding Romanian.
    \item 5 lingual Germanic languages corpus of 1.408M lines which includes da, nl, en, de, and sw.
    \item 4 lingual Uralic languages corpus of 563761 lines which includes en, et, fi and hu.
    \item 3 lingual Baltic languages corpus of 588273 lines which includes lt, li and en.
\end{itemize}

\section{Conclusion}
In this paper we have described our collection of Multilingual Multiway corpora. Our collection, which we believe to be the first of its kind, spans 59 languages and 5 domains. We have also extracted N lingual development and test sets from existing bilingual development and test sets for the IWSLT corpora. For the Bible, Europarl and TED corpora where official  N lingual development and test sets are unavailable we defined our own training, development and test splits and encourage other researchers to use these splits for ease of comparison. Our collection can be used for NLP research in low (Bible and ILCI), medium (IWSLT and TED) as well as high resource (Europarl and UN) scenarios. In the cases of the Bible and Europarl corpora we have extracted N lingual corpora for various language families to facilitate research on how language relatedness affects the final results of NLP tasks. Due to page limit restrictions we do not give various statistics such as word count and average sentence length for each instance of the N lingual corpora we extracted but plan to include it later. We make our collection available to the public\footnote{\url{ http://lotus.kuee.kyoto-u.ac.jp/~raj/mmcr4nlp/}} and also plan on expanding it in the future to improve coverage in terms of domains, number of lines and number of languages. Some additional and promising sources of N lingual corpora are: BTEC corpus\footnote{Although this corpus is large in terms of number of languages and lines it is not easily available} \cite{Paul2013HowTC} and the QED corpus\footnote{http://alt.qcri.org/resources/qedcorpus/}. We also plan on looking into extracting bilingual parallel corpora from Wikipedia and then extract multilingual multiway versions of those corpora.

\section{References}

\bibliographystyle{lrec}
\bibliography{xample}

\end{document}